\documentclass[lettersize,journal]{IEEEtran}
\usepackage{amsmath,amsfonts}
\usepackage{algorithmic}
\usepackage{array}
\usepackage[caption=false,font=normalsize,labelfont=sf,textfont=sf]{subfig}
\usepackage{textcomp}
\usepackage{stfloats}
\usepackage{url}
\usepackage{graphicx}
\usepackage{booktabs}
\usepackage{multirow}
\usepackage{enumitem}
\usepackage{enumerate}
\usepackage[table]{xcolor}
\usepackage{xspace}
\usepackage{makecell}

\usepackage[ruled,vlined]{algorithm2e}
\SetKwInput{KwIn}{Input}
\SetKwInput{KwOut}{Output}

\def\BibTeX{{\rm B\kern-.05em{\sc i\kern-.025em b}\kern-.08em
    T\kern-.1667em\lower.7ex\hbox{E}\kern-.125emX}}
\usepackage{balance}
\begin{document}
\title{Cross Space and Time: A Spatio-Temporal Unitized Model for Traffic Flow Forecasting}

\author{Weilin Ruan, Wenzhuo Wang, Siru Zhong, Wei Chen, Li Liu, and~Yuxuan~Liang
\IEEEcompsocitemizethanks{\IEEEcompsocthanksitem W. Ruan, S. Zhong, W. Chen, and Y. Liang are with the Hong Kong University of Science and Technology (Guangzhou), China.
Corresponding author: Y. Liang (yuxliang@outlook.com)
\IEEEcompsocthanksitem W. Wang is with the Computer Science Department of Jinan University, Guangzhou, China.
\IEEEcompsocthanksitem L. Liu is with the School of Big Data and Software Engineering of Chongqing University, Chongqing, China.
}
}

\maketitle

\def\model{STUM\xspace}

\begin{abstract}
Predicting spatio-temporal traffic flow presents significant challenges due to complex interactions between spatial and temporal factors. Existing approaches often address these dimensions in isolation, neglecting their critical interdependencies. In this paper, we introduce the \underline{S}patio-\underline{T}emporal \underline{U}nitized \underline{M}odel (STUM), a unified framework designed to capture both spatial and temporal dependencies while addressing spatio-temporal heterogeneity through techniques such as distribution alignment and feature fusion. It also ensures both predictive accuracy and computational efficiency. Central to STUM is the Adaptive Spatio-temporal Unitized Cell (ASTUC), which utilizes low-rank matrices to seamlessly store, update, and interact with space, time, as well as their correlations. Our framework is also modular, allowing it to integrate with various spatio-temporal graph neural networks through components such as backbone models, feature extractors, residual fusion blocks, and predictive modules to collectively enhance forecasting outcomes. Experimental results across multiple real-world datasets demonstrate that STUM consistently improves prediction performance with minimal computational cost. These findings are further supported by hyperparameter optimization, pre-training analysis, and result visualization. We provide our source code for reproducibility at 
\url{https://anonymous.4open.science/r/STUM-E4F0}.


\end{abstract}


\section{Introduction}
\noindent 
\IEEEPARstart{R}{apid} economic growth and the surge in vehicle numbers have intensified traffic congestion and parking challenges in urban areas globally. To address these challenges, numerous countries have been investing in the development of Intelligent Transportation Systems (ITS), harnessing advances in data collection and mobile computing technologies~\cite{zhang2011data, zheng2015big, du2018sensable}. Modeling and analyzing spatio-temporal dynamic systems are applicable to various prediction scenarios, and research in this field has received sustained attention over the past few decades~\cite{zhao2019t, zhang2020spatio}. As a crucial component of ITS, traffic flow prediction aims to optimize traffic management, enhance travel safety, and mitigate worsening traffic conditions.~\cite{jin2024survey}

Early research primarily focused on statistical model-based approaches, such as the Historical Average (HA)~\cite{smith1997traffic} and the Auto-Regressive Integrated Moving Average (ARIMA)~\cite{box2015time,lippi2013short} model, as well as machine learning-based models~\cite{van2012short}, including Vector Auto-Regression (VAR)~\cite{lutkepohl2005new,zivot2006vector} and Artificial Neural Networks (ANN)~\cite{huang2014deep}. However, these methods often struggle to capture the complex nonlinear relationships present in large-scale traffic networks, especially when directly applied to spatio-temporal prediction tasks. With the rise of spatio-temporal big data, recent methods have shifted towards data-driven deep learning models that can more effectively capture the inherent spatio-temporal dependencies of dynamic systems~\cite{lv2014traffic}. Simple yet effective strategies include using Convolutional Neural Networks (CNNs)~\cite{gu2018recent} to capture spatial dependencies, and utilizing Recurrent Neural Networks (RNNs)~\cite{yu2019review} and their variants, such as Long Short-Term Memory (LSTM)~\cite{hochreiter1997long}  networks and Gated Recurrent Units (GRUs)~\cite{zhang2018combining}, to capture temporal dependencies, thereby improving performance~\cite{chung2014empirical}.

Recently, numerous traffic prediction methods have combined sophisticated temporal models with Graph Neural Networks (GNNs) to capture global temporal dependencies and regional pattern features, respectively. Spatio-temporal graph neural networks (STGNNs)~\cite{li2017diffusion,yu2017spatio} have gained significant attention due to their ability to learn robust high-level spatio-temporal representations through local information aggregation~\cite{jin2024survey}. Researchers have invested considerable effort in developing complex and innovative models for traffic prediction, including novel graph convolutional methods~\cite{fang2021spatial,guo2021hierarchical,han2021dynamic,wang2020traffic,li2021spatial,ye2021coupled,lu2020spatiotemporal,song2020spatial,wang2023auto,zhao2019t,cao2020spectral,chen2021z,diao2019dynamic}, learning graph structures~\cite{zhang2020spatio,wu2019graph,jiang2023spatio,wu2020connecting,shang2021discrete}, efficient attention mechanisms~\cite{zhou2022fedformer,zhou2021informer,liu2021pyraformer,wu2021autoformer,cirstea2022towards}, and other approaches~\cite{pan2019urban,lee2021learning,pan2021autostg,cirstea2021enhancenet,shao2022pre,zhou2024one,shao2022decoupled}, achieving performance improvements. 

\begin{figure*}[htbp!]
  \vspace{-4mm}
    \centering
    \includegraphics[width=0.95\linewidth]{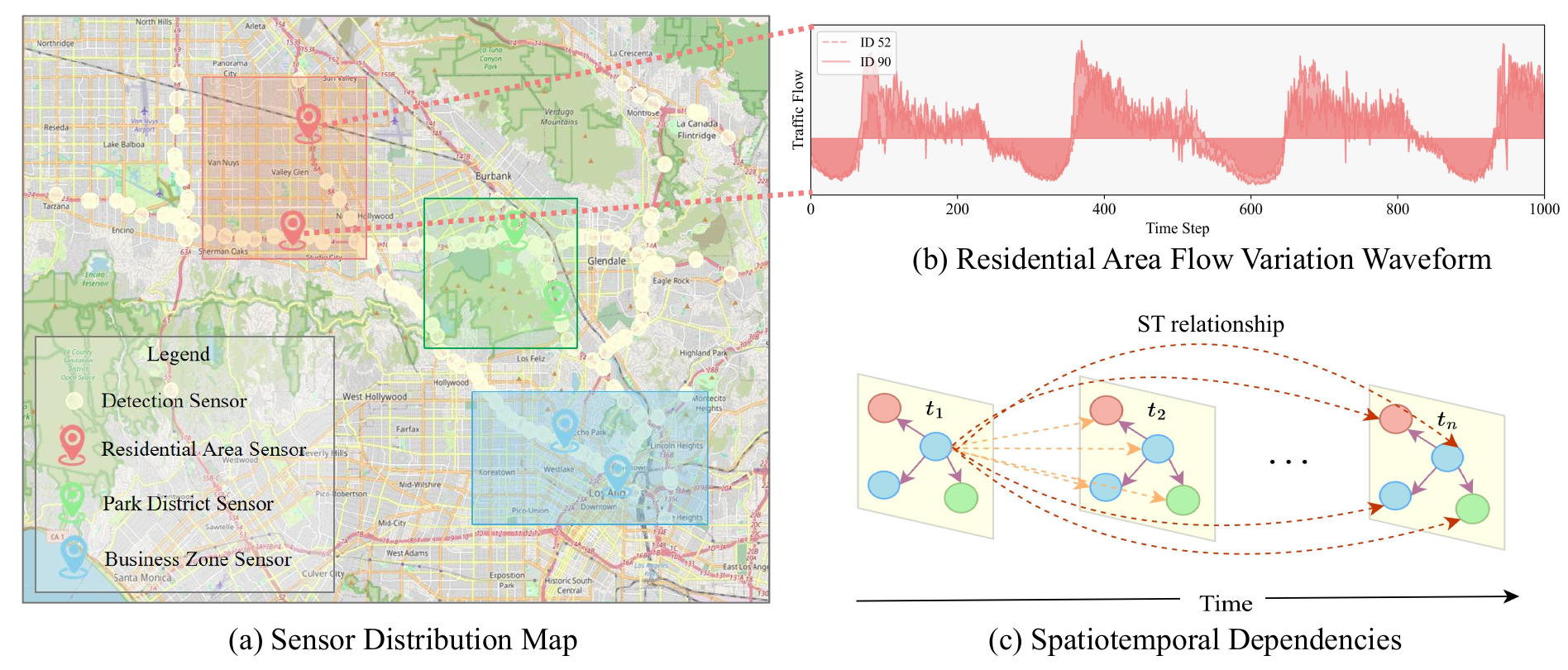}
    \caption{Motivation of our proposed method. (a) shows the sensor distribution of the PEMS04 dataset. (b) is a visual result of the traffic flow of a pair of residential areas over a random period. And (c) is spatio-temporal dependencies shown in traffic flow prediction tasks.}
    \label{fig:motivation}
  \vspace{-6mm}
\end{figure*}

However, despite ongoing advancements in network architectures, performance gains have begun to plateau, largely due to the following challenges:
\begin{itemize}[leftmargin=*]
\item \textbf{Separation between the spatial and temporal module:} The independent computation of spatio-temporal modules always limits the effectiveness and efficiency of spatio-temporal representation learning. As shown in Figure~\ref{fig:motivation}(c), spatio-temporal relational information influences regional predictions over time. Prediction modules that separate spatial and temporal processing fall short of efficiently propagating regional relationships across temporal intervals.

\item \textbf{Data heterogeneity:} The heterogeneity of spatio-temporal data results in varying patterns across different spatial and temporal scales. For instance, Figure~\ref{fig:motivation}(a) depicts one of the regions monitored by sensors in the PEMS dataset~\cite{song2020spatial}, where traffic flow exhibits substantial variability between regions. Figure~\ref{fig:motivation}(b) shows traffic flow waveforms at two points within the same region, highlighting that even within a single area, distinct periods show different traffic dynamics.

\end{itemize}

Upon revisiting existing traffic forecasting methods, we recognize the need for a \textit{unitized framework} to address these challenges. To this end, we first propose the concept of Adaptive Spatio-temporal Unitized Cells (ASTUCs), which are designed to compute, update, and store spatial, temporal, and relational information within a single unit, in contrast to prior research that separates spatial and temporal modules. 
Meanwhile, we propose a novel block called Multi-layer Residual Fusion (MLRF) that leverages the properties of these cells to better capture complex non-linear spatio-temporal dependencies, thereby overcoming heterogeneity and improving computational efficiency and performance. 
Specifically, we begin by defining an adaptive spatio-temporal unitized matrix at the node level, represented by multiple trainable adaptive matrices using low-rank matrix factorization. During the training process, these cells carry node information and aggregate it into reorganized matrices containing dynamic information at each time step. The use of multi-layer fusion residual blocks mitigates redundant computations, reducing over-parameterization. Finally, all adaptive spatio-temporal unitized cells contribute to the prediction module, enabling accurate traffic flow forecasting.

Our main contributions can be summarized as follows:
\begin{itemize}[leftmargin=*]
\item \textit{A unified approach that unifies spatial and temporal learning.} In response to module separation, we introduce a novel framework called the Spatio-temporal Unitized Model (STUM) and a corresponding training approach that unifies spatial and temporal processing, as opposed to the traditional method of separating spatial and temporal modules. This unified treatment allows for more efficient learning and accurate representation of spatio-temporal dependencies.

\item \textit{Designed novel modules for spatio-temporal unitization computing.} In response to data heterogeneity, we present the Adaptive Spatio-temporal Unitized Cell (ASTUC) based on low-rank adaptive matrices, and a dual feature extraction strategy based on backbone network extractor and Multi-layer Residual Fusion (MLRF), improving the model's ability to handle complex spatio-temporal interactions.

\item \textit{Extensive experiments.} We conduct comprehensive experiments on multiple real-world datasets, demonstrating that our proposed framework significantly outperforms existing baseline models in spatio-temporal prediction tasks while maintaining computational efficiency.

\end{itemize}

\begin{figure*}[!t]
    \centering
    \includegraphics[width=0.9\linewidth]{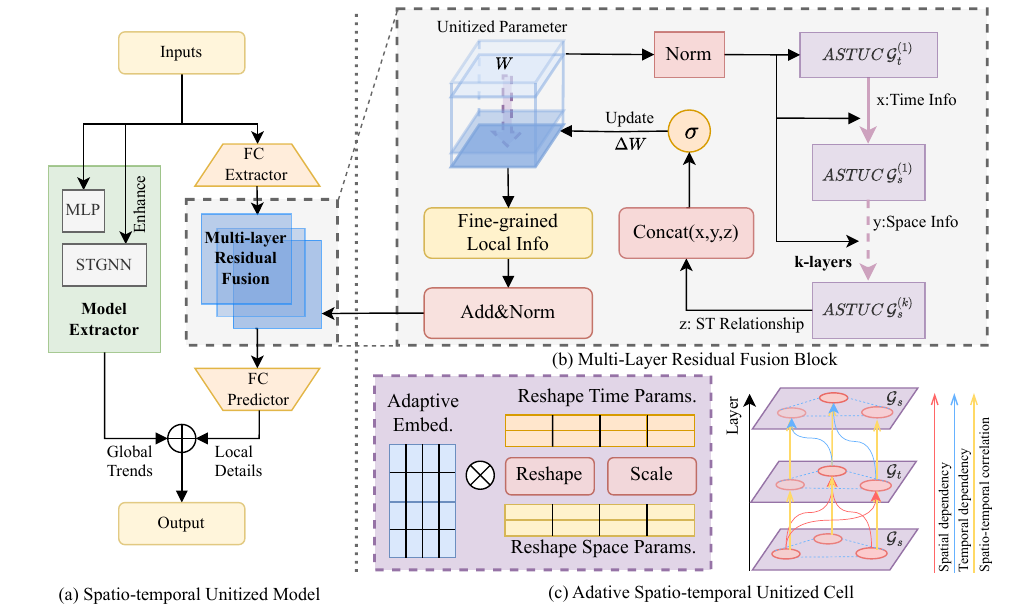}
     \vspace{-1em}
    \caption{The overview of our proposed method. (a) shows the architecture of the Spatio-Temporal Unitized Model (STUM), where MLP represents the model prototype and STGNN represents a way of enhancement. (b) shows the computing process of The Multi-Layer Residual Fusion (MLRF) blocks. (c) shows the construction of Adaptive Spatio-temporal Unitized Cells and how the information transmission Cross Space and Time.}
    \label{fig:overview}
 \vspace{-1em}
\end{figure*}

\section{Related Work}
\subsection{Spatio-temporal Forecasting}
Spatio-temporal forecasting has been extensively studied over the past decades, with the primary objective of predicting future states by analyzing historical data~\cite{zheng2006short,yin2002urban,tan2009aggregation}. Traditional spatio-temporal prediction methods are grounded in statistical methods and time series analysis. While these methods have achieved a certain level of success, they often struggle to effectively capture complex spatial structures and intricate spatio-temporal relationships~\cite{wang2020traffic,jin2023spatio}. To address these challenges, researchers have increasingly turned to deep learning frameworks, which are adept at uncovering sophisticated feature representations, including non-linear spatial and temporal correlations, from historical data~\cite{lv2014traffic, wang2022hierarchical}.

Among these deep learning approaches, Spatio-Temporal Graph Neural Networks (STGNNs) have emerged as particularly powerful tools for prediction tasks. STGNNs integrate Graph Neural Networks (GNNs)~\cite{kipf2016semi} with temporal modeling techniques~\cite{yu2019review}, thereby enhancing their ability to capture complex spatio-temporal dynamics. In recent years, several notable STGNN models have been proposed, including Graph WaveNet~\cite{wu2019graph}, STGCN~\cite{yu2017spatio}, DCRNN~\cite{li2017diffusion}, and AGCRN~\cite{bai2020adaptive}. These models have demonstrated remarkable performance across various spatio-temporal prediction tasks. Additionally, the attention mechanism~\cite{vaswani2017attention} has gained significant popularity due to its effectiveness in modeling dynamic dependencies within spatio-temporal data. Despite the advancements and diversity of STGNN architectures, their performance improvements have begun to plateau. This stagnation has prompted a shift in research focus toward integrating Large Language Models (LLMs)~\cite{zhou2024one, jin2024position, yan2024urbanclip, jin2023time} to further enhance predictive capabilities and overcome existing limitations.

Recent studies have further explored methods to capture and leverage spatio-temporal heterogeneity. Spatial-temporal decoupled masked pre-training~\cite{gao2023spatial} employs separate masked auto-encoders along spatial and temporal dimensions to better learn heterogeneity. Moreover, heterogeneity-informed learning~\cite{dong2024heterogeneity} leverages spatial and temporal embeddings to improve model adaptability and generalization across diverse spatio-temporal contexts. Additionally, self-supervised learning for multi-modality spatio-temporal forecasting~\cite{shao2022pre, li2022spatial, deng2024multimodality, liu2022graph} integrates data augmentation and multi-modality contrastive tasks to enhance the model’s ability to capture heterogeneous patterns across multiple modal domains.

\subsection{Low-Rank Matrix Factorization}
Low-rank matrix factorization aims to decompose high-dimensional matrices into the product of multiple low-rank matrices, thereby reducing computational complexity while minimizing information loss. This approach has wide applications in data compression, dimensionality reduction, missing data recovery, and more~\cite{tan2013tensor}. The deep structure of deep learning models results in numerous training parameters and low training efficiency. Motivated by the idea of data compression, researchers have tried to utilize low-rank matrix factorization to compress deep neural network models to make a trade-off between precision and computational efficiency~\cite{wang2023tensor}. There are mainly two kinds of such compression methods. The first kind compresses the whole deep learning architectures through constructing the corresponding tensor network representation, which has been successfully applied to convolutional architectures~\cite{nekooei2022compression}. The second kind leverages low-rank matrix factorization on the single layers of the network. For instance, ~\cite{calvi2019compression} introduced the Tucker tensor layer as an alternative to the dense weight matrices of neural networks. Recently, the use of low-rank matrix factorization methods in processing graph data has emerged as a lively research field. Low-rank matrix factorization can reveal the hidden information or main components of spatio-temporal data. Graph neural networks (GNNs) perform end-to-end calculations on graph data which contain a vast amount of potential information. To improve the performances of GNNs, researchers have adopted low-rank matrix factorization to mine the hidden information in graph data. For example,~\cite{diao2019dynamic,li2020two,ruan2024low} utilized low-rank matrix factorization to capture spatial and temporal dependencies in traffic data forecasting, thus reducing the computational burden.

\section{Preliminaries}
\subsection{Graph Construction}
A traffic network can be defined as a graph data structure $\mathcal{G} = (V, E, A)$, where $|V| = N$ represents the set of vertices, with $N$ being the number of road segments. Each node corresponds to the location of a road segment, and the observations typically include traffic metrics such as flow and speed on that segment. $E$ represents the set of edges, reflecting the connections between adjacent road segments. The adjacency matrix $A \in \mathbb{R}^{N \times N}$ stores the connectivity information, with each element indicating whether the corresponding road segments are directly connected. Thus, $\mathcal{G}$ captures the spatial relationships between road segments and the spatial dependencies of traffic flow.

\subsection{Problem Definition}
The graph signal matrix is defined as $X^{(t)} \in \mathbb{R}^{N \times C}$, where $C$ is the number of features and $t$ is the time step. $X^{(t)}$ represents the observed values on each road segment at time $t$, such as traffic flow and speed. The traffic prediction task in a sensor network can be formulated as:
\begin{equation}
[X^{(t-s + 1)}, \ldots, X^{(t)}; \mathcal{G}] \stackrel{\mathcal{F}(\cdot),~\theta}{\longrightarrow} [X^{(t+1)}, \ldots, X^{(t+h)}]    
\end{equation}

The input consists of a sequence of graph signals from time step $t-s+1$ to $t$, along with the network structure $\mathcal{G}$. These are mapped to future graph signals from time step $t+1$ to $t+h$ by the learned function $\mathcal{F}(\cdot)$ with parameters $\theta$.

\section{Methodology}
In this section, we introduce a simple yet efficient framework called the Spatio-Temporal Unitized Model (STUM). The overall architecture of STUM is illustrated in Figure~\ref{fig:overview} (a). From input to output, the data is partially processed by the Backbone Extractor to extract global spatio-temporal features, while another portion flows through the Multi-Layer Residual Fusion Blocks (MLRF) as shown in Figure~\ref{fig:overview} (b). The MLRF receives encoded information from a fully connected extractor and computes feature fusion tensors using a fully connected predictor. The Adaptive Spatio-temporal Unitized Cells (ASTUC), implemented using low-rank matrix decomposition as depicted in Figure~\ref{fig:overview} (c), transform the input tensors into time-adaptive and space-adaptive shapes by embedding feature dimensions. By stacking ASTUC modules, STUM captures fine-grained spatio-temporal dependencies, effectively addressing issues of spatio-temporal heterogeneity and the separation of spatial and temporal modules. In the following subsections, we will delve into the core components of STUM and explain the methodology in detail.

\subsection{Dual Feature Extraction of Spatio-temporal Data}
We introduce two key components to extract temporal dependencies and regional correlations from raw data: the backbone network and the adaptive low-rank linear layer. These components are referred to as spatio-temporal feature extractors, denoted as $\mathcal{F}_b$ and $\mathcal{F}_c$, respectively.

Formally, the input sequence is represented as $X = [X_1, \dots, X_s] \in \mathbb{R}^{s \times n \times c}$, where $n$ represents the number of nodes, $s$ is the number of time steps, and each node contains $c$ features such as speed, average flow, and direction. The input $X$ is then mapped into two feature spaces as follows:

\begin{align}
    z_b &= \mathcal{F}_b(X) = [f_1,...,f_h]\in \mathbb{R}^{n \times c_{out}} \\
    X' &= \mathcal{F}_c(X)=[w_1,...,w_h]\in \mathbb{R}^{n\times m}
\end{align}

Here, $f_i$ and $w_i$ represent the global spatio-temporal features extracted by the backbone model and the adaptive spatio-temporal parameters extracted by the $i^{th}$ low-rank fully connected layer, respectively. $c_{\text{out}}$ is the final feature dimension for prediction, and $m$ is the hidden layer dimension.

The backbone network component can be a fundamental module, such as a multi-layer perceptron, which serves as a simple yet essential part of our architectural prototype capable of capturing global spatio-temporal dependencies. Additionally, this component can be replaced with other spatio-temporal graph neural networks (STGNNs) as plug-and-play adapters, offering a flexible means to enhance prediction performance. This process is illustrated in Fig.~\ref{fig:overview}(a).

The low-rank linear layer is designed to reduce redundant weight tensors and computational costs by adopting low-rank matrix decomposition. This approach captures complex spatio-temporal interactions using fewer parameters, making the model more efficient and scalable. Specifically, we let $X'$ be represented as $W^{(i)} = \text{Reshape}(\mathcal{F(\cdot)}) = [w_1^{(i)}, \dots, w_n^{(i)}] \in \mathbb{R}^{N \times M}$ for the $i^{th}$ update iteration, where $N$ and $M$ denote the input and output dimensions. Given a single-step input $x$, the parameter updates and results are computed as:

\begin{align}
\Delta w &= A\times B^T \cdot \frac{r}{\alpha + \epsilon} \\ 
y_i &= \sigma(w^{(i)}x + \Delta w^{(i)}x +b)
\end{align}

Here, $A \in \mathbb{R}^{N \times r}$ and $B \in \mathbb{R}^{M \times r}$ are low-rank matrices, where the intrinsic rank $r \ll \min(N, M)$. $\alpha$ is the scaling factor, and $\sigma$ is the activation function (e.g., ReLU). During inference, $W$ is frozen and does not receive gradient updates, while $A$ and $B$ remain trainable.

\subsection{Unified Representation Cross Space And Time}
The Adaptive Spatio-Temporal Unitized Cell (ASTUC) is a core component designed to simultaneously handle both temporal and spatial information, as well as their interactions within a unified framework. By leveraging low-rank matrix decomposition, ASTUC captures complex spatio-temporal dependencies with fewer learnable parameters, allowing the model to efficiently adapt to specific scenarios without significantly increasing computational complexity. This enables ASTUC to handle spatio-temporal heterogeneity more effectively, enhancing the model's generalization capability.

The key idea behind ASTUC is to compute, update, and store temporal and spatial information in a unified parameter matrix. This matrix is calculated through low-rank matrices, allowing ASTUC to flexibly adapt to the dependencies between different time steps and spatial nodes. The iterative process of passing through ASTUC during the $i^{th}$ update is formalized as follows:
\begin{align}
\mathcal{G}_t^{(i)} &= \text{Update}(X_{:t},\mathcal{G}^{(i-1)}_{s};W,b) \\ 
\mathcal{G}_s^{(i)} &= \text{Update}(X_{:t},\mathcal{G}^{(i-1)}_{t};W,b) \\ 
W \leftarrow \Delta W &= \text{Memory}( \mathcal{G}_t^{(i)}\oplus \mathcal{G}_s^{(i)}, b)
\end{align}

Here, $W$ is the shared parameter matrix that adaptively changes its shape based on the learned spatio-temporal information, while $\oplus$ denotes the joint operation between temporal information $\mathcal{G}_t$ and spatial information $\mathcal{G}_s$. The bias term $b$ is also included. ASTUC alternates between temporal and spatial updates, generating a rich set of interaction information. Given the redundant and low-rank nature of this information, ASTUC selectively retains or forgets specific parts, ensuring that only the most relevant information is passed to the next unitized cell.

\begin{figure}[!t]
  \vspace{-4mm}
  \centering
  \includegraphics[width=0.95\linewidth]{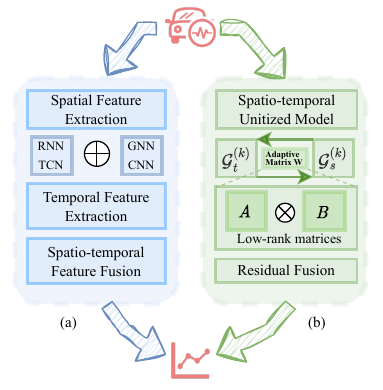}
  \caption{Comparison of traditional and our proposed Methods. The traditional method (a) separates temporal and spatial modeling, while our approach (b) integrates them using low-rank matrix factorization for better joint prediction.}
  \label{fig:comparison}
\end{figure}

By using this unified representation, ASTUC effectively captures dynamic patterns in spatio-temporal data and provides a more comprehensive understanding of the interactions between different time steps and spatial nodes. This process is illustrated in Figure~\ref{fig:comparison}, which compares traditional methods that separate temporal and spatial modeling with our integrated low-rank approach.

\subsection{Global Enhancement and Local Refinement}
The Spatio-Temporal Unitized Model (STUM) is designed not only to achieve strong prediction performance trained from scratch but also to enhance existing spatio-temporal prediction models. This dual capability underscores its high generalization potential. Initially, the backbone extractor captures the global dependencies of spatio-temporal data, providing a comprehensive framework for understanding data structure. This allows STUM to integrate seamlessly with various models, improving their performance through effective global enhancement and local refinement.

Following this, our proposed modules, along with the fully connected extractor and predictor, focus on fine-tuning local details. This means that the backbone extractor can be a fundamental module, such as a multi-layer perceptron (MLP) or convolutional neural network (CNN), or it can utilize presently effective spatio-temporal prediction models like the baselines listed in the experimental part. This flexibility highlights the high generalization and extensibility of our model, allowing for seamless integration with existing methods in a plug-and-play manner.

The integration of the Adaptive Spatio-Temporal Unitized Cell (ASTUC) and the Multi-Layer Residual Fusion block (MLRF) fosters a synergistic effect, enabling the simultaneous extraction of both temporal and spatial features that traditional models often overlook. The modular nature of STUM allows it to be adapted for diverse prediction tasks simply by swapping components, such as employing different backbone networks. This adaptability ensures that STUM can leverage the strengths of various model architectures while maintaining a unified training approach.

Moreover, the residual fusion mechanism within the MLRF block enhances effective information sharing between layers, bolstering the model's capacity to recognize intricate patterns across time and space. By employing a gated mechanism, the prediction outputs from both the backbone network and the MLRF block are dynamically weighted, allowing the model to concentrate on the most relevant information. This results in improved robustness against noise and variability in the input data, ultimately leading to more accurate and reliable predictions.

\subsection{Training and Optimization of STUM Framework}
While a single Adaptive Spatio-Temporal Unitized Cell (ASTUC) can effectively transmit either temporal or spatial information, the complexity and heterogeneity of spatio-temporal data require a more advanced mechanism. To address this, we introduce the Multi-Layer Residual Fusion Block (MLRF), which alternates between the transmission of temporal and spatial information across multiple ASTUCs, enabling joint extraction of temporal, spatial, and spatio-temporal interactions.

The MLRF block is designed to improve spatio-temporal prediction by cross-stacking ASTUCs of different shapes. This design mitigates the typical separation of temporal and spatial encoding found in deep models, offering a unified approach that simultaneously considers temporal dependencies and spatial region patterns.

Each MLRF block first normalizes all parameters and then alternates between spatial and temporal information transmission. We denote the spatial and temporal information passing through the $i^{th}$ spatio-temporal unitized cell as $\mathcal{G}_s^{(i)}$ and $\mathcal{G}_t^{(i)}$. To make the decomposed low-rank matrices better suited to the nonlinear spatio-temporal characteristics, we introduce additional constraints and regularization terms to enhance local adaptability. This allows for a more precise capture of dynamic changes in the data, improving the granularity of feature extraction for prediction. The formal definition throughout this process is as follows:
\begin{align}
    W^{(i)} &= Norm(X) = X\cdot \frac{W^{(i-1)}}{\frac{1}{d}\sum_{i=1}^d x_i^2+\epsilon} \\ 
    \hat{h_i} &= \mathcal{G}_{t}^{(l)}\cdot \mathcal{G}_{s}^{(l)}(\sigma(... \mathcal{G}_{t}^{(1)}\cdot\mathcal{G}_{s}^{(1)}(W^{(i)})...))
\end{align}

Where $X_t \in \mathbb{R}^{n \times f}$ is the spatio-temporal graph input at the $t^{th}$ time step, and $\hat{h_t}$ is the predicted output. The activation function $\sigma$ includes dropout, activation, and regularization terms that control the complexity of the module. By alternating between temporal and spatial information, MLRF captures intricate spatio-temporal patterns at lower computational costs, thereby improving prediction performance.

To address the issue that model parameters may not fully adapt to new or unseen data during training, we propose the Spatio-Temporal Unitized Model (STUM), which combines a spatio-temporal prediction model as the backbone with the plug-and-play MLRF blocks. Finally, we utilize a fully connected layer as the predictor to merge the prediction results from the multi-layer residual fusion module with the predictions of the backbone network. A gated mechanism filters the information to enhance model robustness. The computation process is as follows:
\begin{align}
z_t &= \text{FC}(\sigma(\hat{h_i}))\\
Z &=\mathcal{H}(z_b,z_t;X_{:t},\alpha)=(1-\alpha)\odot \mathcal{F}_b(X_{:t}) + \alpha \odot z_t
\end{align}

Where $\sigma$ denotes the activation function such as Softmax, $\mathcal{F_b}$ is the backbone model’s prediction, $\mathcal{H}$ represents the weighted residual link operator, and $\alpha$ is the update gate coefficient. $Z$ is the final result of spatio-temporal prediction.

\begin{algorithm}[htbp]
\caption{Training and Optimization of STUM}
\KwIn{Spatio-temporal data $\mathbf{X}\in \mathbb{R}^{s\times n\times c_{\text{in}}}$, backbone model $\mathcal{F}_b$, learning rates $\lambda_\theta$, $\lambda_\alpha$}
\KwOut{Trained model parameters $\Theta$, gating coefficient $\alpha$}

Initialize $D_{\text{train}} \leftarrow \emptyset$\;
\For{each $t \in \{1, \dots, T\}$}{
    $X_{\text{input}}\leftarrow(X_{t-\tau_{\text{in}}+1},\dots, X_t)$\;
    $Y_{\text{label}} \leftarrow (X_{t+1},\dots, X_{t+\tau_{\text{out}}})$\;
    Put $\{X_{\text{input}}, Y_{\text{label}}\}$ into $D_{\text{train}}$\;
}

Initialize parameters $\Theta$, $\alpha$\;
\Repeat{stopping criteria is met}{
    Sample $D_{\text{batch}}$ from $D_{\text{train}}$\;
    Compute extraction $z_{b}\leftarrow\mathcal{F}_b(X_{\text{input}})$\;
    $X'\leftarrow\mathcal{F}_c(X_{\text{input}})$\;
    \For{each multi-layer residual fusion $l \in \{1, \dots, L\}$}{
        $\hat{X}^{(l)} \leftarrow \text{RMSNorm}(X_{\text{input}}^{(l-1)})$\;
        $\mathcal{G}_t^{(l)} \leftarrow \text{ASTUC}_{\text{time}}(\hat{X}^{(l)}, \mathcal{G}_s^{(l-1)})$\;
        $\mathcal{G}_s^{(l)} \leftarrow \text{ASTUC}_{\text{space}}(\hat{X}^{(l)}, \mathcal{G}_t^{(l)})$\;
        $\Delta\mathcal{W}^{(l)} \leftarrow \text{Add\&Norm}(\mathcal{G}_t^{(l)}, \mathcal{G}_s^{(l)})$\;
        Update Adaptive Matrix $\mathcal{W} \leftarrow \Delta\mathcal{W}^{(l)}, b$\;
    }
    $z_t \leftarrow \text{Predictor}(\mathcal{W},b)$\;
    $Z \leftarrow (1 - \alpha) \cdot F_{\text{backbone}} + \alpha \cdot z_t$\;
    $L_{\text{train}} \leftarrow \frac{1}{|D_{\text{batch}}|} \sum |Y_{\text{label}} - Z|$\;
    $\Theta \leftarrow \Theta - \lambda_\theta \nabla_\Theta L_{\text{train}}$\;
    $\alpha \leftarrow \alpha - \lambda_\alpha \nabla_\alpha L_{\text{train}}$\;
}

\KwRet{$\Theta$, $\alpha$}
\label{alg:alg1}
\end{algorithm}

Assume the loss function for training the model is $\mathcal{L}_{\text{train}}$, which measures the difference between the predicted values and the ground truth. We can train the STUM model end-to-end using backpropagation. Specifically, there are two types of trainable parameters: the parameters in the feature extraction and prediction layers, denoted as $\Theta$, and the gating coefficient $\alpha$, used for residual fusion. The gradient for the parameters $\Theta$ is $\nabla_{\Theta} \mathcal{L}_{\text{train}}$, while the gradient for the gating coefficient $\alpha$ can be computed using the chain rule, as the residual connection is a differentiable operation:

\[
\nabla_{\alpha} \mathcal{L}_{\text{train}} = \nabla_{Z} \mathcal{L}_{\text{train}} \cdot \nabla_{\alpha} Z
\]


Algorithm~\ref{alg:alg1} outlines the comprehensive training process of the STUM. Initially, the training data is meticulously constructed by organizing the spatio-temporal data into input-output pairs. Following data construction, the STUM framework undergoes iterative optimization using gradient descent-based methods. In each iteration, a batch of training data is sampled, and the model parameters are updated to minimize the prediction error as defined by the chosen loss function.
This iterative training continues until the convergence of the loss function. Throughout the training process, the low-rank matrix factorization within ASTUCs ensures computational efficiency by reducing the number of parameters while preserving essential spatio-temporal relationships. Consequently, the STUM framework achieves a balance between predictive accuracy and computational efficiency, effectively addressing challenges related to data heterogeneity and the separation of spatial and temporal modules.

\section{Experiments}
In this section, we conduct extensive experiments to investigate the following Research Questions (RQ):
\begin{itemize}[leftmargin=*]
\item \textbf{RQ1:} To what extent does our proposed method improve over the baseline models?

\item \textbf{RQ2:} Does the STUM model itself perform well without using STGNN as the global feature extractor?

\item \textbf{RQ3:} What additional training costs in terms of time did we incur to achieve these improvements?
\item \textbf{RQ4:} What differences in results would be observed by using different components or different parameter settings?
\item \textbf{RQ5:} Does our approach truly explain and predict regional traffic flow on a more fine-grained basis?
\end{itemize}

\subsection{Experimental Setup}
\subsubsection{Datasets}
We validate our approach on four real-world datasets widely used in spatio-temporal forecasting. Each dataset comprises tens of thousands of time steps and hundreds of sensors, capturing real-world traffic flow data. Table~\ref{tab:dataset} summarizes the statistical information for each dataset. These datasets were first introduced by~\cite{song2020spatial}. The traffic flow data is represented as integers, with values potentially reaching into the hundreds, reflecting the count of passing vehicles. All datasets are divided into non-overlapping training, validation, and test sets using a 6:2:2 split along the time axis.

\begin{table}[ht]
\centering
\setlength{\tabcolsep}{2.5mm}{}
\caption{Statistics and description of datasets we used.}
\begin{tabular}{c|cccc}
\toprule
Dataset & \#Nodes & \#Edges & \#Frames & Time Range\\
\hline
PEMS03 & 358 & 547 & 26208 & 09/2018 – 11/2018\\
PEMS04 & 307 & 340 & 16992 & 01/2018 – 02/2018\\
PEMS07 & 883 & 866 & 28224 & 05/2017 – 08/2017\\
PEMS08 & 170 & 295 & 17856 & 07/2016 – 08/2016\\
\bottomrule
\end{tabular}
\label{tab:dataset}
\end{table}

\subsubsection{Evaluation Metrics}
We evaluate the performance of our model using three commonly used metrics: Mean Absolute Error (MAE), Root Mean Square Error (RMSE), and Mean Absolute Percentage Error (MAPE). Suppose $x = x_1, ..., x_n$ represents the ground truth, $\hat{x}=\hat{x_1}, ..., \hat{x_n}$ represents the predicted values, and $\Omega$ denotes the indices of observed samples. The metrics are defined as follows:

\begin{align}
MAE(x,\hat{x}) = \frac{1}{|\Omega|}\sum_{i\in \Omega}|x_i - \hat{x_i}|\\
RMSE(x,\hat{x}) = \sqrt{\frac{1}{|\Omega|}\sum_{i\in \Omega}(x_i - \hat{x_i})^2}\\
MAPE(x,\hat{x}) = \frac{1}{|\Omega|}\sum_{i\in \Omega}|\frac{x_i - \hat{x_i}}{x_i}|
\end{align}

\subsubsection{Baselines}
We compare the performance of our proposed approach with the following traffic flow prediction models. The core mechanisms of the following baseline models are summarized:
\begin{itemize}[leftmargin=*]
\item \textbf{STGCN.} Spatio-Temporal Graph Convolution Network~\cite{yu2017spatio}, which combines graph convolutions with 1D temporal convolutions to jointly model spatial and temporal dependencies.
\item \textbf{GWNet.} Graph WaveNet~\cite{wu2019graph}, which enhances spatial correlation modeling with an adaptive adjacency matrix and employs 1D dilated convolutions to capture temporal dependencies in traffic data.
\item \textbf{AGCRN.} Adaptive Graph Convolutional Recurrent Network~\cite{bai2020adaptive}, which introduces the Node Adaptive Parameter Learning (NAPL) module and Data Adaptive Graph Generation (DAGG) to automatically infer interdependencies in traffic flow time series.
\item \textbf{D2STGNN.} Decoupled Dynamic Spatio-Temporal Graph Neural Network~\cite{shao2022decoupled}, which decouples dynamic graph learning by separating diffusion and inherent traffic information, improving the model’s ability to capture hidden temporal signals.
\item \textbf{STAE.} Spatio-Temporal Adaptive Embedding Transformer~\cite{liu2023spatio}, which leverages a linear layer to expand feature dimensions and applies several embedding layers to encode node, spatial, and temporal characteristics separately.
\item \textbf{STID.} Spatial-Temporal Identity~\cite{shao2022spatial}, a lightweight MLP-based model that attaches spatial and temporal identity embeddings to capture the uniqueness of each sample in multivariate time series (MTS) forecasting, achieving competitive results without complex graph structures.
\end{itemize}

\subsubsection{Implementation Details}
We implement the model with the PyTorch toolkit on a Linux server with NVIDIA RTX A6000 GPUs. The training process utilizes the Adam optimizer, with an initial learning rate set to 0.001 and a weight decay of 0.0005 for regularization. We train each model for a maximum of 150 epochs, with early stopping applied if the validation loss does not improve for 10 consecutive epochs. The batch size is set to 64. For the final results, we select the average performance of all predicted 12 horizons on the PEMS03, PEMS04, PEMS07, and PEMS08 datasets. We perform significance test (t-test with p-value $\leq$ 0.05) over all the experimental results. For any other more details, readers could refer to our public code repository.

\subsection{Performance Comparisons (RQ1 and RQ2)}
\begin{table*}[htbp!]
  \centering
  \caption{Overall prediction performance of different methods on the PEMS03,04,07,08 datasets, results with $\Delta$ are reported improvement of our STUM model with corresponding backbone extractor compared to the original model. And a smaller metric value means better performance.}
  \resizebox{0.95\textwidth}{!}{
    \begin{tabular}{c|c|c|c|c|c|c|c|c|c|c|c|c|c}
    \toprule
    \multirow{2}{*}{\rotatebox{90}{\textbf{ }}} & \multirow{2}{*}{Model} & \multicolumn{3}{c|}{PEMS03} & \multicolumn{3}{c|}{PEMS04} & \multicolumn{3}{c|}{PEMS07} & \multicolumn{3}{c}{PEMS08} \\
    \cmidrule{3-14} 
          &  & MAE↓  & RMSE↓ & MAPE↓ & MAE↓  & RMSE↓ & MAPE↓ & MAE↓  & RMSE↓ & MAPE↓ & MAE↓  & RMSE↓ & MAPE↓ \\
    \midrule
    \multirow{18}{*}{\rotatebox{90}{\textbf{STUM Enhancement}}} 
        & STGCN & 17.27  & 28.72  & 17.74\% & 20.62  & 31.98  & 15.27\% & 24.21  & 37.38  & 11.31\% & 16.58  & 25.65  & 11.27\% \\
        & Ours & 15.42  & 24.10  & 15.48\% & 19.75  & 30.85  & 14.84\% & 23.56  & 36.66  & 10.67\% & 15.80  & 25.38  & 10.52\% \\
        & $\Delta$ & \cellcolor{green!20} -1.85  & \cellcolor{green!20} -4.62  & \cellcolor{green!20} -2.26\% & \cellcolor{green!20} -0.87  & \cellcolor{green!20} -1.12  & \cellcolor{green!20} -0.43\% & \cellcolor{green!20} -0.65  & \cellcolor{green!20} -0.72  & \cellcolor{green!20} -0.64\% & \cellcolor{green!20} -0.78  & \cellcolor{green!20} -0.26  & \cellcolor{green!20} -0.75\% \\
        \cline{2-14}
        & GWNet & 15.16  & 25.82  & 16.11\% & 19.88  & 31.37  & 13.96\% & 22.52  & 35.97  & 9.69\% & 14.92  & 23.76  & 9.89\% \\
        & Ours & 14.91  & 24.96  & 15.83\% & 19.32  & 30.72  & 13.60\% & 21.99  & 35.33  & 9.41\% & 14.86  & 23.69  & 9.77\% \\
        & $\Delta$  & \cellcolor{green!20} -0.25  & \cellcolor{green!20} -0.86  & \cellcolor{green!20} -0.28\% & \cellcolor{green!20} -0.56  & \cellcolor{green!20} -0.65  & \cellcolor{green!20} -0.36\% & \cellcolor{green!20} -0.54  & \cellcolor{green!20} -0.65  & \cellcolor{green!20} -0.28\% & \cellcolor{green!20} -0.06  & \cellcolor{green!20} -0.08  & \cellcolor{green!20} -0.12\% \\
        \cline{2-14}
        & AGCRN & 16.69  & 27.60  & 16.44\% & 20.74  & 32.61  & 14.57\% & 23.29  & 36.18  & 10.07\% & 15.30  & 24.51  & 10.29\% \\
        & Ours & 15.49  & 26.79  & 15.58\% & 19.03  & 30.67  & 13.41\% & 22.20  & 35.05  & 9.80\% & 15.25  & 24.27  & 10.11\% \\
        & $\Delta$  & \cellcolor{green!20} -1.20  & \cellcolor{green!20} -0.81  & \cellcolor{green!20} -0.86\% & \cellcolor{green!20} -1.71  & \cellcolor{green!20} -1.94  & \cellcolor{green!20} -1.16\% & \cellcolor{green!20} -1.10  & \cellcolor{green!20} -1.13  & \cellcolor{green!20} -0.27\% & \cellcolor{green!20} -0.05  & \cellcolor{green!20} -0.24  & \cellcolor{green!20} -0.18\% \\
        \cline{2-14}
        & STAE  & 15.29  & 25.87  & 17.64\% & 20.59  & 32.71  & 14.79\% & 21.97  & 34.81  & 9.86\% & 14.71  & 23.79  & 10.15\% \\
        & Ours & 15.23  & 25.45  & 16.63\% & 18.93  & 30.32  & 13.27\% & 21.57  & 34.40  & \multicolumn{1}{c}{9.64\%} & 14.62  & \multicolumn{1}{c}{23.65} & 10.11\% \\
        & $\Delta$  & \cellcolor{green!20} -0.06  & \cellcolor{green!20} -0.42  & \cellcolor{green!20} -1.01\% & \cellcolor{green!20} -1.66  & \cellcolor{green!20} -2.39  & \cellcolor{green!20} -1.52\% & \cellcolor{green!20} -0.40  & \cellcolor{green!20} -0.40  & \cellcolor{green!20} -0.22\% & \cellcolor{green!20} -0.09  & \cellcolor{green!20} -0.14  & \cellcolor{green!20} -0.04\% \\
        \cline{2-14}
        & STID  & 15.33  & 27.40  & 16.40\% & 19.58  & 31.79  & 13.38\% & 21.52  & 36.29  & 9.15\% & 15.58  & 25.89  & 10.33\% \\
        & STUM+STID & 15.26  & 25.77  & 16.37\% & 18.55  & 29.95  & 12.85\% & 19.99  & 32.96  & \multicolumn{1}{c}{8.58\%} & 14.51  & 23.44  & 9.45\% \\
        & $\Delta$  & \cellcolor{green!20} -0.07  & \cellcolor{green!20} -1.63  & \cellcolor{green!20} -0.03\% & \cellcolor{green!20} -1.03  & \cellcolor{green!20} -1.84  & \cellcolor{green!20} -0.53\% & \cellcolor{green!20} -1.53  & \cellcolor{green!20} -3.33  & \cellcolor{green!20} -0.57\% & \cellcolor{green!20} -1.07  & \cellcolor{green!20} -2.45  & \cellcolor{green!20} -0.88\% \\
        \cline{2-14}
        & D2STGNN & 15.76  & 26.45  & 14.89\% & 22.85  & 35.23  & 17.33\% & 21.20  & 34.09  & 9.18\% & 15.72  & 24.67  & 11.46\% \\
        & Ours & 15.24  & 26.10  & 16.00\% & 21.16  & 33.05  & 15.08\% & 20.79  & 33.67  & \multicolumn{1}{c}{9.04\%} & 15.67  & 24.64  & 11.32\% \\
        & $\Delta$  & \cellcolor{green!20} -0.52  & \cellcolor{green!20} -0.36  & \cellcolor{green!20} 1.11\% & \cellcolor{green!20} -1.70  & \cellcolor{green!20} -2.18  & \cellcolor{green!20} -2.25\% & \cellcolor{green!20} -0.41  & \cellcolor{green!20} -0.41  & \cellcolor{green!20} -0.14\% & \cellcolor{green!20} -0.04  & \cellcolor{green!20} -0.04  & \cellcolor{green!20} -0.14\% \\
    \bottomrule
    \end{tabular}%
  }
  \label{tab:performance}%
\end{table*}%

\begin{table}
\caption{Comparison of STGNNs and STUM Framework without enhancement (We use MLP as a backbone extractor). {\color{black}$\mathbf{H}$ denotes horizon}. Numbers marked with $^*$ indicate that the improvement is statistically significant compared with the best baseline~(t-test with p-value$<0.05$).}
\label{tab:result_vanilla}
\resizebox{0.48\textwidth}{!}{
\begin{tabular}{p{0.1cm}<{\centering}p{0.24cm}<{\centering}p{1cm}<{\centering}p{1cm}<{\centering}<{\centering}p{1cm}<{\centering}p{1cm}<{\centering}p{1cm}<{\centering}}
\toprule
  \midrule 
& & & STGCN & GWNet & AGCRN  & STUM\\
      \midrule
\multirow{9}*{\rotatebox{90}{\textbf{PEMS03}}} 
&\multirow{3}*{\makecell{\textbf{H}\\\textbf{3}}}     
& MAE  & 15.98 & 13.74 & 14.41 & \textbf{13.63}$^*$   \\ 
& & RMSE  & 26.67 & 23.35 & 25.03 & \textbf{23.00}$^*$    \\ 
& & MAPE  & 17.44\% & 14.62\% & 15.19\% & \textbf{14.04\%}$^*$  \\ 
\cmidrule(r){2-7}
&\multirow{3}*{\makecell{\textbf{H}\\\textbf{6}}}     
& MAE  & 17.00 & 15.07 & 15.62 & \textbf{14.89}$^*$   \\ 
& & RMSE  & 28.54 & 25.65 & 27.21 & \textbf{25.25}$^*$    \\ 
& & MAPE  & 17.96\% & 16.25\% & 15.82\% & \textbf{15.34\%}$^*$  \\ 
\cmidrule(r){2-7}
&\multirow{3}*{\makecell{\textbf{H}\\\textbf{12}}}    
& MAE  & 19.29 & 17.28 & 17.38 & \textbf{17.05}$^*$   \\ 
& & RMSE  & 32.09 & 29.01 & 30.08 & \textbf{28.54}$^*$    \\ 
& & MAPE  & 20.12\% & 17.57\% & 17.89\% & \textbf{17.20\%}$^*$  \\ 
\midrule
\midrule
\multirow{9}*{\rotatebox{90}{\textbf{PEMS04}}} 
&\multirow{3}*{\makecell{\textbf{H}\\\textbf{3}}}     
& MAE  & 19.69 & 18.52 & 18.24 & \textbf{18.15}$^*$   \\ 
& & RMSE  & 30.69 & 29.54 & 29.54 & \textbf{29.36}$^*$    \\ 
& & MAPE  & 14.27\% & 12.84\% & 12.75\% & \textbf{12.71\%}$^*$  \\ 
\cmidrule(r){2-7}
&\multirow{3}*{\makecell{\textbf{H}\\\textbf{6}}}     
& MAE  & 20.64 & 19.84 & 19.07 & \textbf{18.96}$^*$   \\ 
& & RMSE  & 32.28 & 31.38 & 31.09 & \textbf{30.87}$^*$    \\ 
& & MAPE  & 14.84\% & 13.88\% & 13.33\% & \textbf{13.17\%}$^*$  \\ 
\cmidrule(r){2-7}
&\multirow{3}*{\makecell{\textbf{H}\\\textbf{12}}}    
& MAE  & 22.34 & 22.05 & 20.30 & \textbf{20.15}$^*$   \\ 
& & RMSE  & 34.89 & 34.28 & 32.97 & \textbf{32.74}$^*$    \\ 
& & MAPE  & 15.87\% & 15.89\% & 14.32\% & \textbf{14.24\%}$^*$  \\ 
\midrule
\midrule
\multirow{9}*{\rotatebox{90}{\textbf{PEMS07}}}
&\multirow{3}*{\makecell{\textbf{H}\\\textbf{3}}}     
& MAE  & 22.63 & 19.68 & 19.57 & \textbf{19.41}$^*$   \\ 
& & RMSE  & 34.61 & 31.85 & 31.40 & \textbf{31.26}$^*$    \\ 
& & MAPE  & 10.61\% & \textbf{8.42\%}$^*$ & 8.52\% & 8.57\% \\ 
\cmidrule(r){2-7}
&\multirow{3}*{\makecell{\textbf{H}\\\textbf{6}}}     
& MAE  & 24.22 & 21.82 & 20.93 & \textbf{20.75}$^*$   \\ 
& & RMSE  & 37.32 & 35.28 & 34.02 & \textbf{33.88}$^*$    \\ 
& & MAPE  & 11.17\% & 9.31\% & 8.90\% & \textbf{8.90\%}$^*$  \\ 
\cmidrule(r){2-7}
&\multirow{3}*{\makecell{\textbf{H}\\\textbf{12}}} 
& MAE  & 27.09 & 25.48 & 23.02 & \textbf{22.79}$^*$   \\ 
& & RMSE  & 41.85 & 40.57 & 37.59 & \textbf{37.39}$^*$    \\ 
& & MAPE  & 12.21\% & 11.12\% & 10.14\% & \textbf{10.05\%}$^*$  \\ 
\midrule
\midrule
\multirow{9}*{\rotatebox{90}{\textbf{PEMS08}}}
&\multirow{3}*{\makecell{\textbf{H}\\\textbf{3}}}     
& MAE  & 15.78 & 14.02 & 14.41 & \textbf{13.88}$^*$   \\ 
& & RMSE  & 24.04 & 22.14 & 22.65 & \textbf{22.00}$^*$    \\ 
& & MAPE  & 11.21\% & 9.05\% & 9.72\% & \textbf{8.84\%}$^*$  \\ 
\cmidrule(r){2-7}
&\multirow{3}*{\makecell{\textbf{H}\\\textbf{6}}}     
& MAE  & 16.57 & 15.03 & 15.34 & \textbf{14.86}$^*$   \\ 
& & RMSE  & 25.66 & 24.00 & 24.61 & \textbf{23.82}$^*$    \\ 
& & MAPE  & 11.40\% & 9.90\% & 10.27\% & \textbf{9.63\%}$^*$  \\ 
\cmidrule(r){2-7}
&\multirow{3}*{\makecell{\textbf{H}\\\textbf{12}}}    
& MAE  & 18.23 & 16.79 & 16.67 & \textbf{16.51}$^*$   \\ 
& & RMSE  & 28.29 & 26.61 & 27.11 & \textbf{26.35}$^*$    \\ 
& & MAPE  & 12.41\% & 11.25\% & \textbf{11.04\%}$^*$  & 11.24\% \\ 

\midrule  
\bottomrule
\end{tabular}
}
\end{table}

Each baseline model has been widely used in spatio-temporal forecasting and offers a distinct approach to handling spatial and temporal dependencies. We use these baseline models as backbone extractors to improve the performance of STUM across various metrics. As shown in Table~\ref{tab:performance}, all methods trained as backbone network feature extractors combined with the STUM framework achieved more optimal performance than the original model in all datasets, which indicates that our model is very effective. 
STGCN shows the most significant improvement (about 19.17\%) when combined with STUM. This is likely due to the limited ability to capture complex spatio-temporal dependencies, which separate temporal and spatial convolutions. GWN still struggles to fully capture intricate temporal patterns in rapidly changing traffic conditions. While STUM equipped GWN enabling it to capture finer temporal patterns and regional interactions more effectively (about 5.47\% improvement). Despite this, AGCRN and D2STGNN, which are more advanced models with adaptive mechanisms for learning spatial dependencies, also benefit (about 8.99\% and 8.03\%) from the addition of STUM. While their original performance is already strong, STUM further enhances their ability to capture dynamic spatio-temporal relationships due to fine-grained local information. STAE, as a SOTA model in recent years, improved (about 8.77\%) from STUM relatively smaller compared to other models because it already incorporates sophisticated mechanisms for encoding spatio-temporal interactions. Nevertheless, the addition of STUM refines the model’s ability to fine-tune regional and temporal interactions, resulting in a modest but consistent enhancement in overall performance. STID has a relatively simple structure relying on simple identity embeddings. Due to this nature, STID can only achieve a relatively small 6.32\% enhancement when used in combination with our method.

To further examine the independent performance of STUM without relying on other Spatio-temporal graph neural networks as a backbone extractor to capture global features, we conducted additional experiments across the PEMS03, PEMS04, PEMS07, and PEMS08 datasets. As presented in Table~\ref{tab:result_vanilla}, the results indicate that STUM performs robustly even without advanced global feature extraction. Across both short-term and long-term forecasting tasks, STUM consistently outperforms the three baseline models including STGCN, GWNet and ACGRN in almost all cases, demonstrating its ability to capture local spatio-temporal patterns effectively. This highlights that even without complex global feature extraction, STUM exhibits strong standalone performance, making it a viable and efficient model for spatio-temporal forecasting tasks.

\subsection{Efficiency Analysis (RQ3)}
\begin{figure}[htbp!]
    \centering
    \includegraphics[width=.95\linewidth]{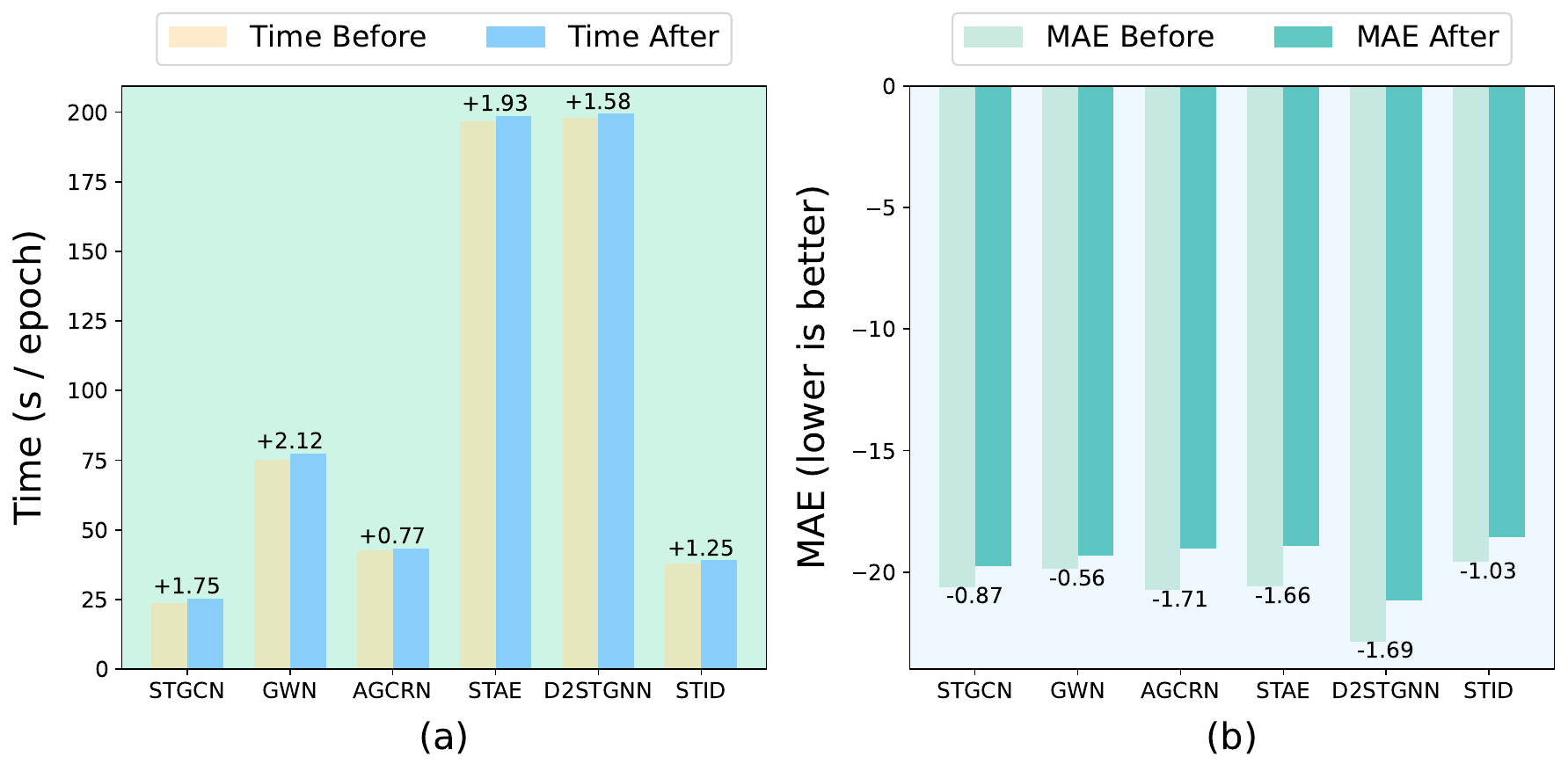}
    \caption{The Efficiency Study. The results compare the variation in training time and the reduction in MAE metrics for STUM equipped with all six baselines as backbone networks on the PEMS04 dataset.}
    \label{fig:efficiency}
\end{figure}

We significantly increased the effectiveness of the model with only a small amount of additional training cost. Figure~\ref{fig:efficiency} (a) shows the difference in time cost between training some models from scratch before and after combining them with our proposed framework, while Figure~\ref{fig:efficiency} (b) illustrates the reduction in MAE metrics. 
We fixed the number of MLRFs to 4, the number of ASTUCs $\mathcal{G}_s$ and $\mathcal{G}_t$ to 8, and the embedding dimension to 16. We observe that the low-rank adaptation portion of our framework allows the training time to remain stable even when multiple ASTUCs are used. These improvements are achieved with minimal additional training time, highlighting the efficiency of our framework in balancing both accuracy and computational cost.

\subsection{Ablation Study (RQ4)}
In this section, we conduct a comprehensive ablation study to analyze the sensitivity of various hyper-parameters in our model and the impact of different feature extractors. Specifically, we varied the number of MLRFs, ASTUCs, and the embedding dimension while maintaining other settings consistent with RQ2. Table~\ref{tab:parameter_sensitivity} summarizes the results, elucidating the trade-offs between model complexity and prediction accuracy.

\begin{table}[htbp]
  \centering
  \caption{Parameter Sensitivity Analysis on PEMS04 Dataset. The table shows the effect on the long-term forecasting task of each module and varying the number of MLRFs, ASTUCs, as well as embedding dimensions in the STUM framework.}
    \resizebox{0.48\textwidth}{!}{
    \begin{tabular}{lrrr}
    \toprule
    Method & MAE & RMSE & MAPE \\
    \midrule
    Compared Baseline(AGCRN) & 25.09  & 37.97  & 19.56\% \\
    STUM~(Default)  & 1.28$\downarrow$   & 1.50$\downarrow$   & 2.15\%$\downarrow$ \\
    w/o Backbone~(use MLP) & 0.39$\downarrow$   & 0.21$\downarrow$   & 1.87\%$\downarrow$ \\
    w/o MLRF\&ASTUC & 0.36$\downarrow$   & 0.20$\downarrow$   & 1.06\%$\downarrow$ \\
    \midrule
    STUM (MLRFs=5) & 1.21$\downarrow$   & 1.40$\downarrow$   & 2.47\%$\downarrow$ \\
    STUM (MLRFs=6) & 1.31$\downarrow$   & 1.47$\downarrow$   & 2.50\%$\downarrow$ \\
    STUM (MLRFs=7) & 1.40$\downarrow$   & 1.70$\downarrow$   & 2.15\%$\downarrow$ \\
    STUM (MLRFs=8) & 1.37$\downarrow$   & 1.58$\downarrow$   & 2.58\%$\downarrow$ \\
    \midrule
    STUM (ASTUCs=10) & 1.40$\downarrow$   & 1.61$\downarrow$   & 2.79\%$\downarrow$ \\
    STUM (ASTUCs=12) & 1.54$\downarrow$   & 1.80$\downarrow$   & 2.19\%$\downarrow$ \\
    STUM (ASTUCs=14) & 1.59$\downarrow$   & 1.82$\downarrow$   & 2.37\%$\downarrow$ \\
    STUM (ASTUCs=16) & 1.62$\downarrow$   & 1.88$\downarrow$   & 2.81\%$\downarrow$ \\
    \midrule
    STUM (Embed\_dims=12) & 1.16$\downarrow$   & 1.35$\downarrow$   & 1.81\%$\downarrow$ \\
    STUM (Embed\_dims=16)  & 1.28$\downarrow$   & 1.50$\downarrow$   & 2.15\%$\downarrow$ \\
    STUM (Embed\_dims=20) & 1.57$\downarrow$   & 1.62$\downarrow$   & 2.65\%$\downarrow$ \\
    STUM (Embed\_dims=24) & 1.69$\downarrow$   & 1.75$\downarrow$   & 2.90\%$\downarrow$ \\
    STUM (Embed\_dims=28) & 1.62$\downarrow$   & 1.62$\downarrow$   & 2.95\%$\downarrow$ \\
    \bottomrule
    \end{tabular}
}
  \label{tab:parameter_sensitivity}
\end{table}
The results demonstrate that integrating STUM with AGCRN provides significant improvements in long-term (60mins) spatio-temporal forecasting. Each component of the STUM framework plays a vital role. Removing either the backbone extractor or our proposed modules (MLRFs and ASTUCs) significantly diminishes the optimization effect. However, even when replacing the backbone extractor with a simpler MLP, the model still achieves meaningful improvements, indicating that our framework does not heavily rely on the specific forecasting model, thus showcasing strong generalization capabilities.

Moreover, when we increased the number of MLRFs, ASTUCs, or the embedding dimension while keeping other parameters fixed, the model's performance consistently improved, confirming that all these parameters contribute positively to the model’s predictive ability. Among these, increasing the number of ASTUCs provided the largest gain. However, increasing the embedding dimension or excessively adding residual fusion layers led to diminishing returns. This is because deeper residual fusion modules are prone to gradient vanishing and exploding problems, which can degrade performance. Additionally, increasing the number of parameters also makes training more difficult. The embedding dimension reflects the intrinsic rank of the parameter matrix, and an excessively large intrinsic rank can make the model overly complex and challenging to learn.

The low-rank matrix factorization mechanism in our framework ensures that the additional computational cost resulting from increasing the number of ASTUC layers is minimized, while significantly enhancing the model's ability to capture more intricate spatio-temporal dependencies. However, to achieve optimal performance, it is advisable to adjust other parameters alongside increasing the ASTUC layers to balance the model's complexity and accuracy.

\subsection{Visualization Case Study (RQ5)}

To further illustrate why STUM is effective, we present a case study on enhancing region embeddings learned from the MLRF. This visualization highlights two key features optimized by the STUM framework: 1) Region refinement, where neighboring regions with similar traffic flows are closely clustered; and 2) Spatio-temporal utilization, where uniform nodes from different time steps are closer together in the embedding space.

\begin{figure}[htbp]
\vspace{-4mm}
    \centering
    \includegraphics[width=.98\linewidth]{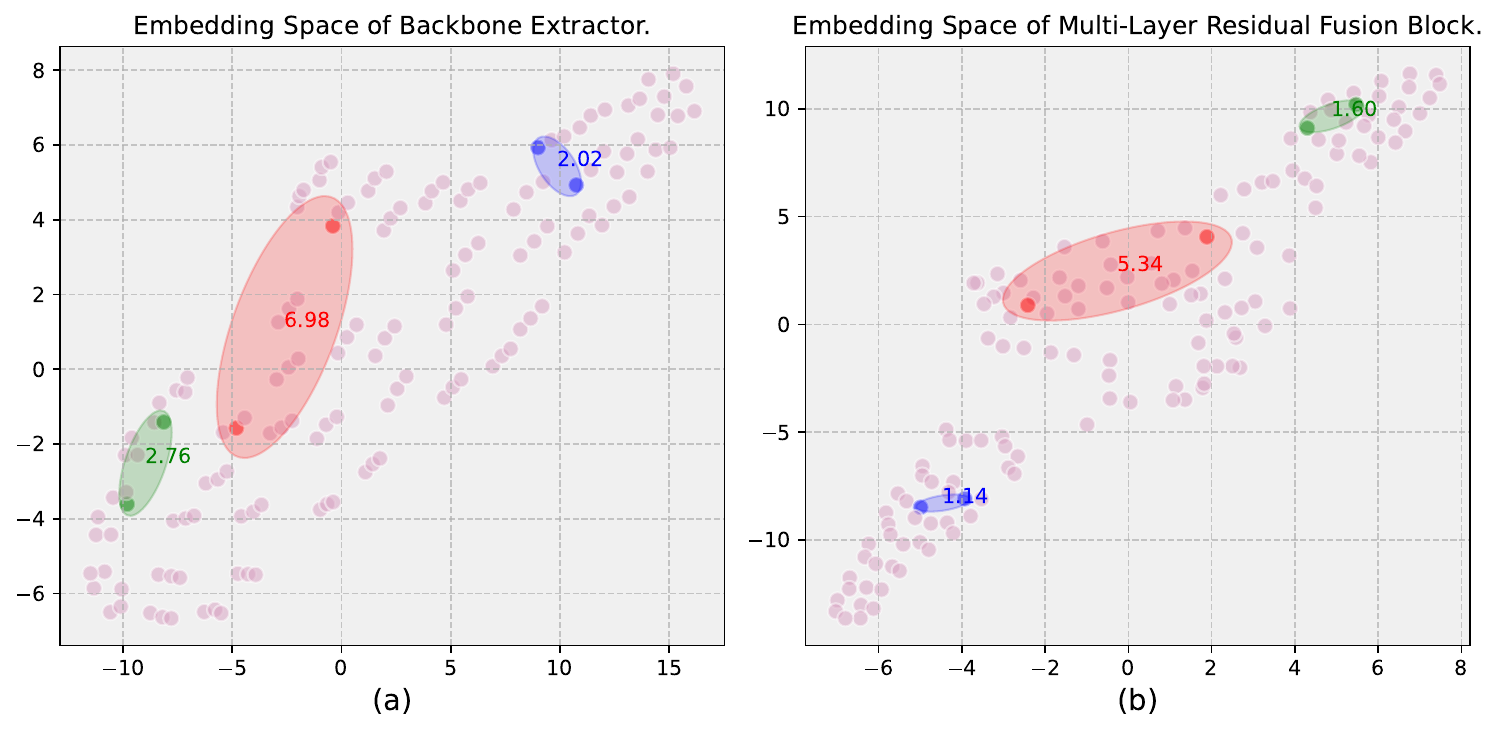}
    \caption{Results of the t-SNE visualization of the Spatio-temporal Unitized Model embedding on the PEMS04 dataset. The left part represents the embedding space obtained using STAEFormer as the backbone extractor, and the right side represents the embedding space enhanced by Multi-Layer Residual Fusion equipped with four Adaptive Spatio-temporal Unitized Cells.}
    \label{fig:tsne}
\end{figure}

As shown in Figure~\ref{fig:tsne} (a), we use the t-SNE node embedding visualization in the 2D plane to demonstrate that the backbone extractor has captured the dispersed traffic patterns on the PEMS04 dataset. In Figure~\ref{fig:tsne} (b), the MLRF further clusters regional information more precisely, thus showing that the STUM framework provides a more fine-grained interpretation and prediction of regional traffic flow. Noted that we have highlighted three pairs of points representing different regions corresponding to Figure~\ref{fig:motivation}: red points 52 and 90 represent residential areas, green points 37 and 61 represent park districts, and blue points 93 and 114 represent business districts. The distances labeled between each pair of points intuitively demonstrate how our method effectively clusters and refines regional information, providing clearer distinctions and insights. Compared to other models, the results show that our method better gathers the sensor points with similar characteristics, thereby clustering their spatio-temporal information. This validates both the effectiveness and generalization capability of our module.

\begin{figure}[htbp]
    \centering
    \includegraphics[width=.98\linewidth]{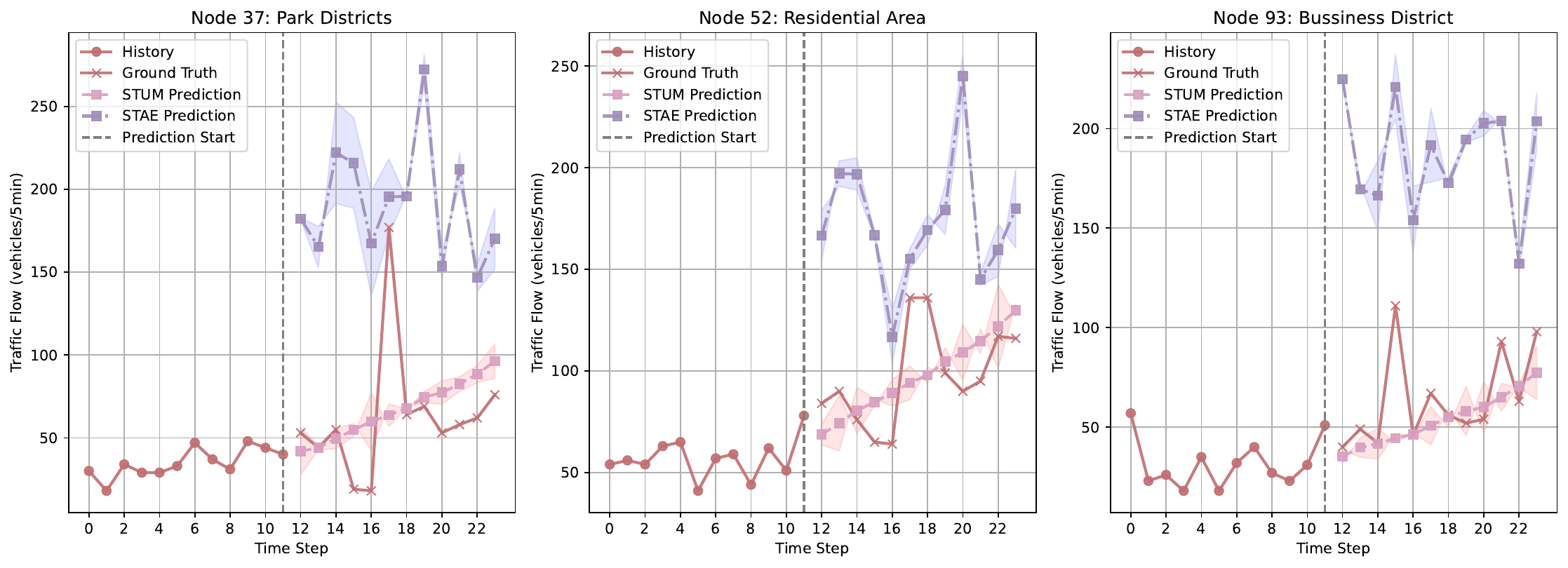}
    \caption{Visualizations of the traffic flow prediction on the PEMS04 dataset. The comparison of STAEFormer highlights the prediction effect of our model.}
    \label{fig:prediction}
\end{figure}
In the visualization of traffic flow prediction results shown in Figure~\ref{fig:prediction}, the STUM model consistently outperforms the baseline models across various regions, particularly excelling at capturing traffic flow trends. Specifically, in the three nodes representing a park district, residential area, and business district, the STUM model accurately predicts the downward trends in traffic flow. In regions with more volatile traffic patterns (e.g., the business district), its predictions are closer to the ground truth compared to the STAE model. Furthermore, the STUM model consistently aligns well with actual data during the early stages of traffic flow decline, demonstrating its robust capability in modeling complex spatio-temporal features. This superior predictive performance further confirms the effectiveness and advantages of the STUM model in spatio-temporal traffic forecasting tasks.

\section{Conclusion}
In this paper, we address several critical challenges in spatio-temporal forecasting, including data heterogeneity, the separation of spatial and temporal modules, and low combination efficiency. To tackle these issues, we introduce the STUM, which unifies spatial and temporal processing in a single framework. Our approach leverages the ASTUCs to effectively capture complex spatio-temporal dependencies. This unified module directly addresses the inefficiency caused by spatial and temporal module separation, ensuring that region relationships are propagated more effectively across different time steps. Furthermore, by incorporating multi-layer residual fusion modules, we mitigate the computational burden and improve combination efficiency without compromising performance. Extensive experiments and analyses demonstrate that our approach consistently outperforms existing methods across various benchmarks. In the future, we plan to explore the potential of a unified spatio-temporal framework for multi-task generalization while integrating additional optimization techniques to further boost efficiency and performance.

\section{Acknowledgments}

This work is mainly supported by the National Natural Science Foundation of China (No.
62402414). This work is also supported by the Guangzhou-HKUST(GZ) Joint Funding Program (No. 2024A03J0620), Guangzhou Municipal Science and Technology Project (No. 2023A03J0011), the Guangzhou Industrial Information and Intelligent Key Laboratory Project (No. 2024A03J0628), and a grant from State Key Laboratory of Resources and Environmental Information System, and Guangdong Provincial Key Lab of Integrated Communication, Sensing and Computation for Ubiquitous Internet of Things (No. 2023B1212010007).

\bibliographystyle{IEEEtran}
\bibliography{IEEEabrv.bib}

\end{document}